\pdfoutput=1
\documentclass[conference, letterpaper]{IEEEtran}
\newcommand{\RNum}[1]{\uppercase\expandafter{\romannumeral #1\relax}}

\ifCLASSINFOpdf
\else
\fi
\usepackage{subcaption}

\ifCLASSINFOpdf
   \usepackage[pdftex]{graphicx}
\else
\fi

\usepackage[cmex10]{amsmath}
\usepackage{color}
\usepackage{fancyhdr}

\renewcommand{\thispagestyle}[2]{}

\fancypagestyle{plain}{
        \fancyhead{}
        \fancyhead[C]{first page center header}
        \fancyfoot{}
        \fancyfoot[C]{first page center footer}
}
\usepackage{siunitx}
\newcommand{\rad}[1]{\SI{#1}{\radian}}
\usepackage{epstopdf}
\begin{document}

%
\title{End to End Deep Neural Network Frequency Demodulation of Speech Signals}

\author{\IEEEauthorblockN{Dan Elbaz}
\IEEEauthorblockA{Department of Computer Science\\
Technion Israel Institute of Technology\\
32000 Haifa, Israel\\
Email: elbazdan@gmail.com}
\and

\IEEEauthorblockN{Michael Zibulevsky}
\IEEEauthorblockA{Department of Computer Science\\
Technion Israel Institute of Technology\\
32000 Haifa, Israel}}


%


\maketitle

\begin{abstract}
Frequency modulation (FM) is a form of radio broadcasting which is widely used nowadays and has been for almost a century. We suggest a software-defined-radio (SDR) receiver for FM demodulation that adopts an end-to-end learning based approach and utilizes the prior information of transmitted speech message in the demodulation process. The receiver detects and enhances speech from the in-phase and quadrature components of its base band version. The new system yields high performance detection for both acoustical disturbances, and communication channel noise and is foreseen to out-perform the established methods for low signal to noise ratio (SNR) conditions in both mean square error and in perceptual evaluation of speech quality score.
\end{abstract}



%
\IEEEpeerreviewmaketitle

\section{Introduction}
Frequency modulation (FM) is a nonlinear encoding of information on a carrier wave. It can be used for interferometric, seismic prospecting, telemetry and many more applications, each with its own statistics, dominated by the underling generating process. However, its widest use is for radio broadcasting, which is commonly used for transmitting audio signal representing voice.\\
Communication transmission channel is subject to various distortions, noise conditions and other impairments. Those impairments severely degrade FM demodulator performance when a critical level is exceeded. As a result thereof, the intelligibility and quality of the detected speech decreases significantly. This phenomenon is known as the Threshold Effect.\\
Long Short-Term Memory (LSTM) recurrent neural networks \cite{lstm} are powerful models that can capture long range dependencies and non-linear dynamics. LSTMs achieve excellent performance on a general sequence to sequence learning problems \cite{sutskever}, and are used for solving many difficult problems such as text to speech mapping \cite{fan} language translation and speech and music generation ( \cite{zen}, \cite{gruv}).\\ 

This paper introduces FM demodulator based on Long Short-Term Memory recurrent neural network.
The main contributions of this work are as follows:
\begin{itemize}
\item utilizing the LSTM abilities to capture the temporal dynamics of speech signals and take advantage of the prior statistics of the speech to overcome transmission channel disturbances.
\end{itemize}
\begin{itemize}
\item Taking an end-to-end learning based approach for filtering both acoustical disturbances, modeled as phase noise and transmission channel disturbances, modeled as amplitude noise. In this approach, the LSTM learn to map directly from the modulated baseband signal to the modulating audio that had been applied at the transmitter, thus creating a baseband to speech mapping.
\end{itemize}
We demonstrate this method by applying it to Frequency modulation (FM) decoding in varying levels of amplitude and phase noise and show it has a superior performance over legacy reception systems in low SNR conditions.

\section{Problem formulation and related work}
Traditionally radio transmission decoding and speech enhancement are considered as two separate problems. However, optimal signal estimation algorithms are usually constructed on the basis of statistical properties of a measurement process and prior statistical or deterministic model of the reconstructed signals. This is often a difficult problem with no analytic solution that can be only approximately solved based on simplistic models of noise and signal. On the other hand, any signal estimation can be considered as a non-linear mapping from input data to the desired output. Having a universal function approximation tool in hand, we can learn such mapping using a set of training examples, pairs of input modulated baseband signals and the desired audio output signals.\\
Except for the traditional methods for radio transmission decoding and for minimum mean square error (MMSE) based speech enhancement techniques, several, though not many, neural networks based methods have been proposed for each of the two problems separately.
For example the radio transmission decoding: \cite{digidemod1}, \cite{digidemod2} and recently \cite{digidemod3}, and for channel noise estimation \cite{digidemod4}, however, these works deal with digital communication for which bit-streams are mapped to symbols, moreover, traditionally the symbols are precoded and scrambled before transmitted, therefore effectively the coded data stream is uncorrelated from time-sample to time-sample  \cite{commbook}, and use of the prior speech data to overcome the noise in the transmission channel is not possible.\\
The fact that the modulating input is proportional only to the instantaneous frequency of the received FM signal has driven the development of traditional FM demodulators to rely on very short time frame processing in order to extract the modulating signal, disregarding long range dependencies that are present in the transmitted voice.\\
\cite{anademod} addressed the analog FM problem, but the approach taken was to imitate the way a conventional FM demodulator works by implementing different neural network for each building block separately. It used memoryless (or very short memory) feed-forward neural network with only one input at some intermediate blocks, and therefore it did not take into account the prior knowledge of the transmitted speech. Moreover, demodulation was performed directly on the passband signal so the input of the neural network needed to be sampled with very high sampling rate in order to detect the change in the input, resulting in several samples, most of which are redundant, and a very large network for actual sampling rates that is very difficult to train and not suitable for practical use.\\
As for the problem of speech enhancement, several neural networks based solutions were suggested and shown to give good performance, for example \cite{speech1}, \cite{speech2}, \cite{speech3} and in \cite{speechlstm} an LSTM based model was suggested.\\
While the listed works have applied neural networks to the task of radio demodulation and for speech enhancement separately, neither of these works suggested the task of radio transmission decoding with the prior information of transmitted speech messages. In this sense, our project is entirely novel as our network exploits the prior knowledge of the speech signal to overcome both acoustical disturbances and noise in the communication channel and performs audio reconstruction by directly operating on the baseband representation of the modulated data.\\ 
In the work, Demodulation as Probabilistic Inference \cite{prob}, demodulation is viewed as a problem of inference and learning and it was suggested to use a demodulation process that can be shaped by user-specific prior information. We adopt this approach in our work and suggest a neural network based solution for this problem.

\section{Background}
\subsection{Signal modulation}
Frequency Modulation is the process of modulating a sine wave with an information message \(x_m\left ( t \right )\) in the following manner:
\[y\left ( t \right )= A_{c} cos\Big(2\pi f_{c} t+2\pi f_{\Delta }\int_{0}^{t}x_{m}\left (  \tau \right )d\tau\Big)\]
Where \(x_{m}\left ( t \right )\) is the data signal, which is typically a speech signal, \(x_{c}(t)=A_{c}\cos(2\pi f_{c}t)\) is the sinusoidal carrier, \(f_{c}\)  is the base frequency of the carrier, \(A_{c}\) is the amplitude of the carrier and \(f_{\Delta }\) is the frequency deviation, which represents the maximum shift away he carrier's base frequency.\\

\subsubsection{Noise Model}
During the process of transmission and reception, the signal is subject to several impairments. On the receiving side, the modulator's role is to reconstruct the original signal from the received signal with maximal level of reliability, overcoming the impairments introduced by the transmission and reception phases.\\
As mentioned, the message signal undergoes several distortions, the signal impairments due to those distortions can be divided into two categories:
\begin{enumerate}
  \item Phase noise: Impairments due to environmental conditions such as audio distortions and the operation of frequency modulation, those original audio additive impairments are translated to the phase to become phase noise. 
 
\noindent
\begin{equation*}
  r\left ( t \right )= A_{c} cos\Big(2\pi f_{c} t+2\pi f_{\Delta }\int_{0}^{t}\left (  x_{m}\left (  \tau \right )+n\left (  \tau\right )\right )d\tau\Big)\\
\end{equation*}

    \item Amplitude noise: Impairments due to communication channel distortions such as convolution with the communication channel, multi-path, additive noise due to propagation characteristics of the channel environment, etc. those impairments are translated to additive amplitude noise, \(r\left ( t \right )=y\left ( t \right )+n\left ( t \right )\).\\
\end{enumerate}
In communication systems, the statistical model for each of the above noise models is usually assumed to be white Gaussian noise.
For clarity a Fig. 1 presents a diagram depicting the communication system and its elements.
\begin{figure}[!t]
\centering
\includegraphics[width=0.4\textwidth]{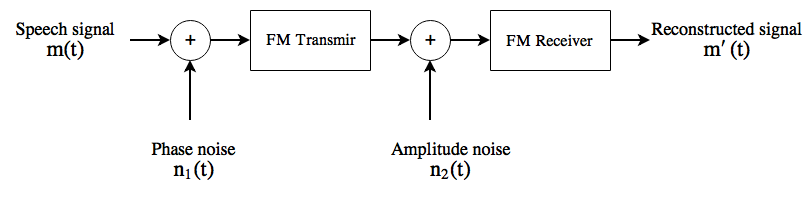}
\caption{Communication system with amplitude and phase noise sources.}
\label{fig1}
\end{figure}
\subsubsection{Baseband equivalents of bandpass
signals}
Sinusoid with frequency modulation can be decomposed into two amplitude-modulated sinusoids that are offset in phase by one-quarter cycle (\rad{\pi/2}). The amplitude modulated sinusoids are known as in-phase and quadrature components or the I/Q components.
By using simple trigonometric identities the general expression representing the transmitted signal can be expressed as follows:
\begin{align*}
\noindent
 	& y\left ( t \right ) = \\
	& A_c cos\left ( 2\pi f_ct \right )cos\left ( 2\pi f_{\Delta }\int_{0}^{t}x_m\left ( \tau  \right )d\tau \right ) - \\
	&A_c sin\left ( 2\pi f_ct \right )sin\left ( 2\pi f_{\Delta }\int_{0}^{t}x_m\left ( \tau  \right )d\tau  \right ) \\
\end{align*}
The I/Q components can be defined in the following way:
\begin{align*}
       &I\left ( t \right )=A_c cos\left ( 2\pi f_{\Delta }\int_{0}^{t}x_m\left ( \tau  \right )d\tau  \right )\\
       &Q\left ( t \right ) = A_c sin\left ( 2\pi f_{\Delta }\int_{0}^{t}x_m\left ( \tau  \right )d\tau  \right )\\
\end{align*}
and we can represent the modulated signal with its I/Q components in the following way:
\[y\left ( t \right )=I\left ( t \right )cos\left (2\pi f_ct \right )-Q\left ( t \right ) sin\left ( 2\pi f_ct \right )\]
This signal has a bandpass spectrum centered around the carrier frequency \(f_{c}\).\\
It is common to analyze communication systems by using the low pass equivalents, also referred to as baseband (or I/Q components) of the original band pass signals.\\

\subsection{Long Short-Term Memory Networks}
Long short-term memory (LSTM) is a recurrent neural network architecture.
Due to the vanishing and exploding gradient \cite{Pascanu:2013:DTR:3042817.3043083} the training of recurrent neural networks is a challenging task. To address this issue, the LSTM cell has been introduced by \cite{lstm}. We are using the common LSTM version \cite{graves13}, with the following update equations:
\begin{align*}
       & i_{t}=\sigma_{i}\left ( W_{xi}x_{t}+W_{hi}h_{t-1}+W_{ci}c_{t-1}+b_{i}\right )\\
       & f_{t}=\sigma_{f}\left ( W_{xf}x_{t}+W_{hf}h_{t-1}+W_{cf}c_{t-1}+b_{f}\right )\\
       & c_{t}=f_{t}c_{t-1}+i_{t}tanh\left ( W_{xc}x_{t}+W_{hc}h_{t-1}+b_{c} \right )\\
       & o_{t}=\sigma_{o}\left ( W_{xo}x_{t}+W_{ho}h_{t-1}+W_{co}c_{t}+b_{o}\right )\\
       & h_{t}=o_{t}tanh\left ( c_{t} \right )                          
\end{align*}
where \(\sigma\) is the logistic sigmoid function, \(i, f, o\) and \(c\) are respectively the input gate, forget gate, output gate and cell activation vector cells at time \(t\). \(x_{t}\) is the input feature vector, \(h_{t}\) is hidden output vector,
\(b_{i}, b_{f}\) and \(b_{o}\) are the bias terms and
 \(W_{hi}, W_{hf},W_{ho},W_{xi},W_{xf}\) and \(W_{xo}\) are the weight matrices connecting the different inputs and gates with the memory cells.

\subsection{Natural speech structure}
Natural speech is composed of many timescale features, generated by anatomic processes that control sound production.
A typical segment of speech can be decomposed to sentences or words that are of a typical time scale of one second. On a smaller time scale, words can be decomposed into phonemes, which are one of the units of sound  that distinguish one word from another. Usually phonemes last a duration which is smaller than \(10^{-1}\) seconds. We can look on an even smaller time scale, such as pitch \(10^{-2}\) and formants \(10^{-3}\). For an optimal reconstruction to take place, all those timescales need to be accounted for in the reconstruction task.\\

\section{Method description}
\subsection{Dataset and training procedure }
In order to support high quality audio transmissions broadcast, FM stations use large values of frequency deviation. The FM broadcast standards in the United States specify a value of 75kHz of peak deviation and 240kHz sampling frequency of the output signal. The default value of the modulating audio signal is 48kHz.
For the above reasons, the training set was generated using Matlab FM modulation \cite{matlab} with the above stated standard specifications.
The above system constraints dictates the number of baseband samples the modulator produces (five in-phase and five quadrature) for each audio sample on its input.
In order to avoid manipulating FM passband signal directly, we assume that conversion to baseband from intermediate frequency will be performed by another digital or analog hardware. This conversion process is known as synchronous detection or heterodyning the signal down to baseband and it is usually performed in the analog front end.
Converting the high frequency signal to baseband signal, enables more convenient processing in a lower sampling rate than the original carrier frequency and alleviates the demodulator (either standard or DNN based) computational demands.
The audio waveforms used in our experiments were downloaded from TIMIT Acoustic-Phonetic Continuous Speech Corpus \cite{timit}. The TIMIT corpus includes 16-bit, 16kHz speech waveform file for each utterance. We used male speakers from New England dialect region. 
The speech material in the TIMIT corpus is subdivided into portions for training and
testing. The criteria for the subdivision has no relation to the data distributed, and can by found in the corpus documentation.  
For the input of the neural network we used two features, samples of the in-phase and quadrature components of the baseband signal.
For compatibility with standard United States specifications described above the waveforms were up-sampled to 48 kHz.

\subsection{Architecture}
In many signal estimation tasks, the advantage of recurrent neural network becomes significant only when there is a statistical dependency between the examples. We utilize those abilities of the recurrent neural network, more specifically LSTM, to capture the temporal dynamics of speech signals for the problem of source signal estimation from noisy frequency-modulated measurements.
Since there is a direct mapping between the generating speech and the modulated FM signal, future baseband samples are also related to previous baseband samples, as dictated by the underling generating speech. We would like the demodulator to exploit future context as well as prior speech statistics in the speech demodulation task. 
To exploit this dependency we introduce a small delay of 100 samples, this enables us to use bidirectional RNNs \cite{brnn}, this network is trained using  input information from the past as well as from the future of a specific time frame.
This is achieved by processing the data in both directions with two separate hidden layers.
For higher level representations of the modulated speech signal we use deep architectures. Deep RNNs can be created by stacking multiple LSTM layers on top of each other, with the output sequence of one layer forming the input sequence for the next. The stacking of multiple recurrent hidden layers can combine the multiple levels or representations of speech with flexible use of long range context, and have proven to give state-of-the-art performance for acoustic modeling \cite{stack}, \cite{stack2}. 
The fact that our inputs (baseband samples) and outputs (audio samples) are synced sequences, and for the above stated reasons we have decided to adopt Deep bidirectional LSTM architecture based on the architecture proposed in \cite{stack}. For regularization we added a dropout layer \cite{dropout}.
We trained with batches of 512 examples and truncated backpropagation through time to length of 100 time steps. Using the system specification described in Dataset and training procedure section dictates ten baseband samples for each audio sample, i.e. in training we unrolled the network to 100 time steps, in each time step the input is a baseband vector of length ten, and the output is its corresponding audio sample, As illustrated in Fig. 2.
Long term dependencies were accounted for by preserving the network state between batches, as the last state of a batch was used as the initial state for the following batch.
The entire system was optimised with  RmsProp\cite{rmsprop} optimization method, using backpropagation through time \cite{Rumelhart:1988:LRB:65669.104451}.
\begin{figure}[!t]
\centering
\includegraphics[width=0.4\textwidth]{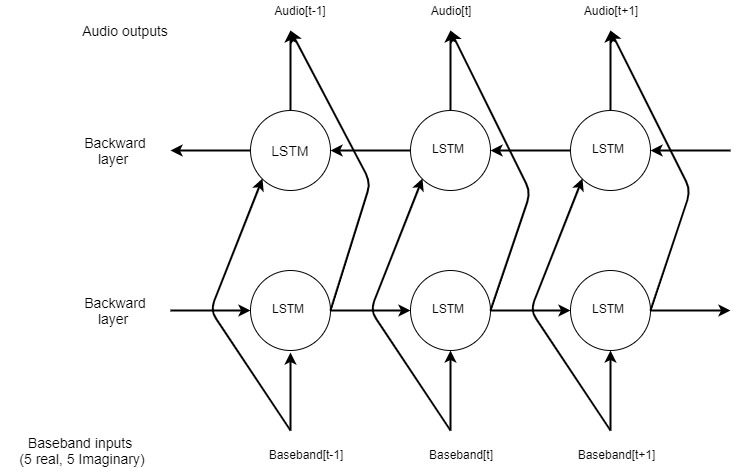}
\caption{Network architecture, stacked bidirectional LSTM}
\label{fig2}
\end{figure}
\section{Experimental results}
In order to evaluate the performance of the DNN demodulator we used both the Mean Squared Error (MSE) objective measure and Perceptual Evaluation of Speech Quality, PESQ\cite{pesq}. PESQ was particularly developed to model subjective tests commonly used in telecommunications and is a more suitable measure for speech quality assessment. The aim of the PESQ score is to achieve high correlation with majority opinion score (MOS) tests for speech quality assessment as perceived by human beings. 
We compare the performance of proposed LSTM demodulator against the performance of conventional demodulator implementation from Matlab communication toolbox, which is based on \cite{matlab}. In both cases, DNN and conventional demodulator, the modulated signal sample rate is set to 240 kHz and the frequency deviation is set to 75 kHz (United States standard).
In order to boost the performance of the conventional FM receiver and compensate FM characteristic in that it amplifies high frequency noise and degrades the overall signal-to-noise ratio, we used matlab FM broadcasters. FM broadcasters insert a pre-emphasis filter prior to FM modulation to amplify the high-frequency content. The FM receiver has a reciprocal de-emphasis filter after the FM demodulator to attenuate high-frequency noise and restore a flat signal spectrum. For clarity the full FM broadcast system is depicted in figure 3. 
\begin{figure}[!t]
\centering
\includegraphics[width=0.4\textwidth]{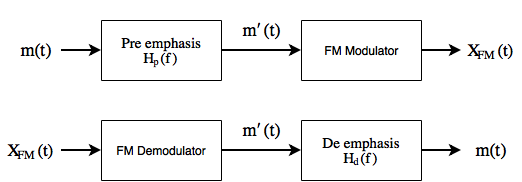}
\caption{Full broadcast system, used for performance comparison}
\label{fig3}
\end{figure}
We start with the noise free case, i.e. neither phase nor amplitude noise were added to the modulated signal. For the noise free case we get output SNR of 36.56 dB and the PESQ score measured between the reconstructed and the original audio is 4.54, these results indicate high quality reconstruction.
Fig. 4 shows the spectrogram of the original audio and spectrogram of the LSTM demodulator reconstruction for noise free case.
\begin{figure}[!t]
\centering
\includegraphics[width=0.4\textwidth]{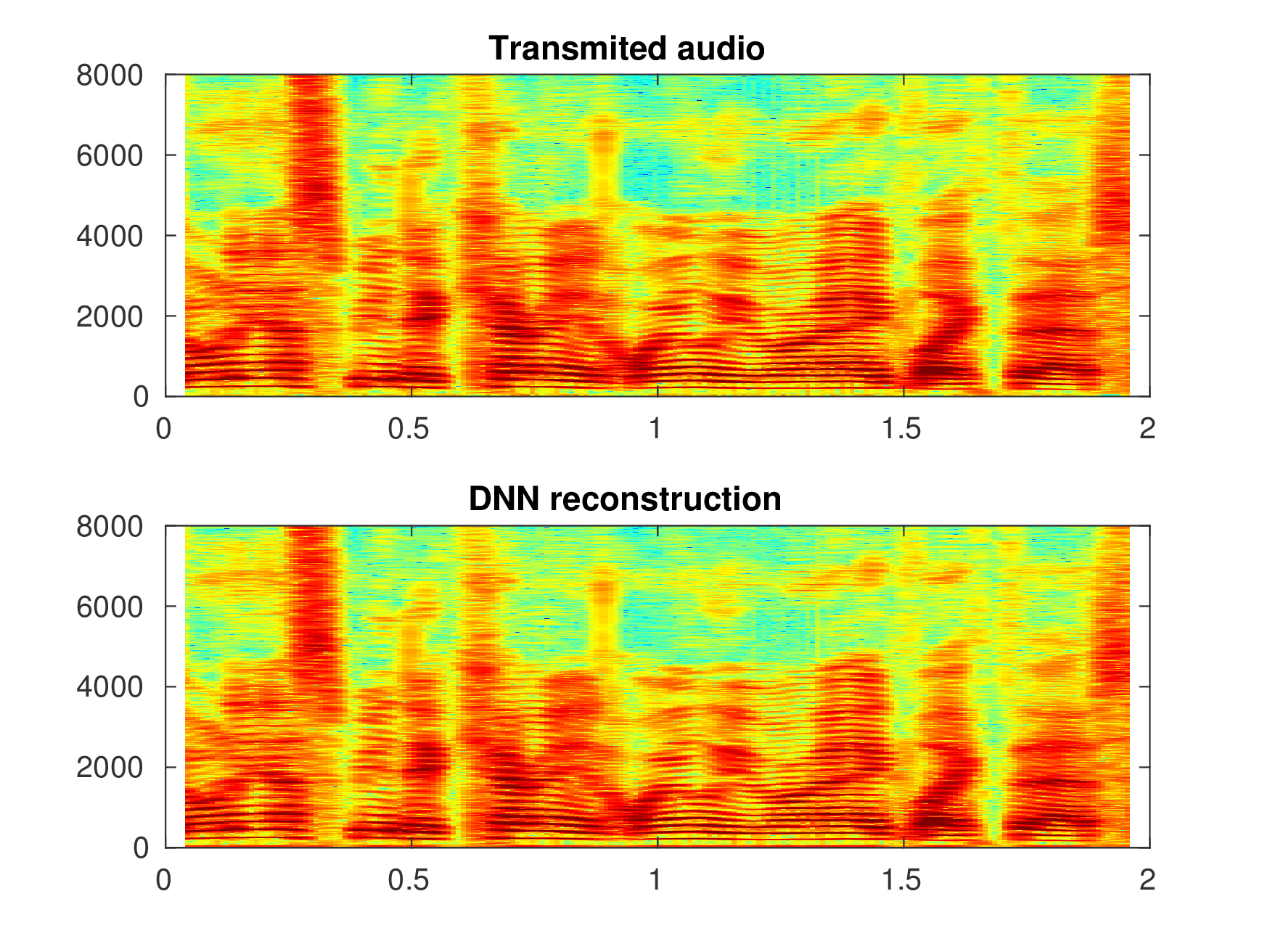}
\caption{Spectrogram of the original audio signal and DNN demodulator reconstruction.}
\label{fig4}
\end{figure}
We investigated the performance of the proposed scheme, by testing the reconstruction quality in experiments employing FM modulation for various levels of additive white Gaussian (AWGN) amplitude noise. 
Fig. 5 shows MSE of the speech reconstruction comparison between the proposed LSTM demodulator and conventional
demodulator for various levels of AWGN amplitude noise.
Fig. 6 shows PESQ score of the speech reconstruction comparison between the proposed LSTM demodulator and conventional
demodulator for various levels of AWGN amplitude noise.
\begin{figure}[!t]
\centering
\includegraphics[width=0.4\textwidth]{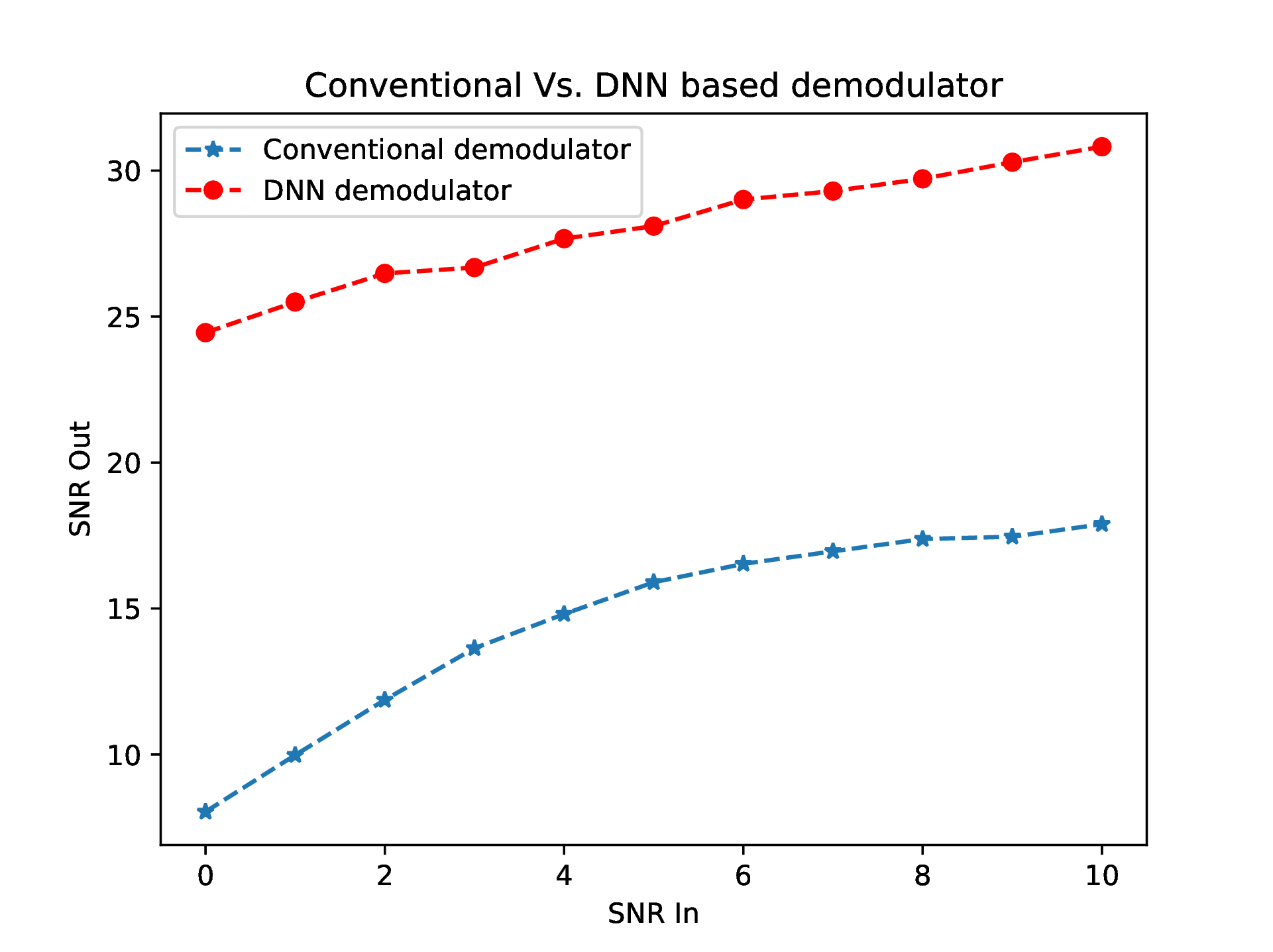}
\caption{Output SNR of the audio reconstruction Vs. channel input SNR}
\label{fig5}
\end{figure}

\begin{figure}[!t]
\centering
\includegraphics[width=0.4\textwidth]{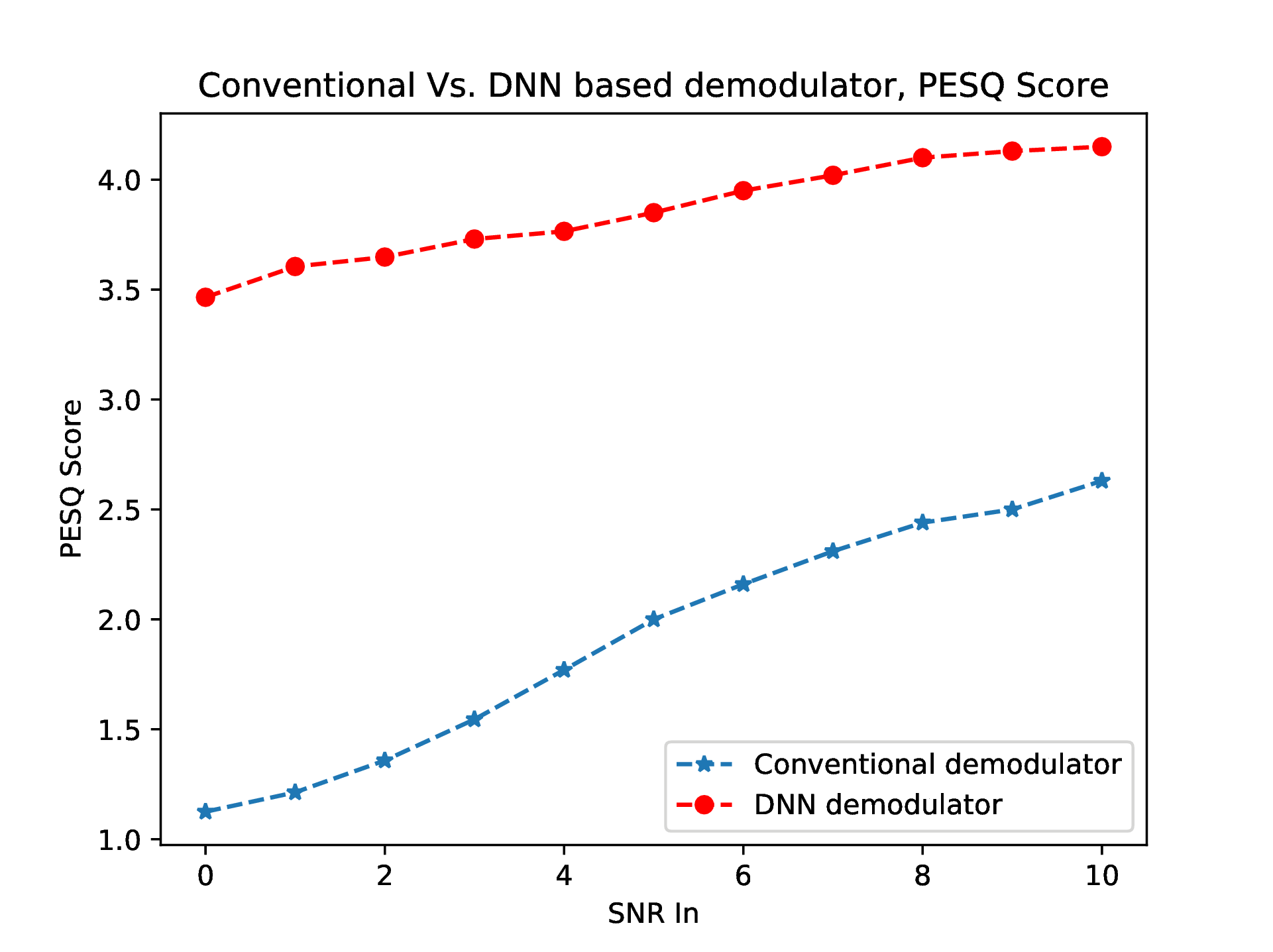}
\caption{PESQ score of the audio reconstruction Vs. channel input SNR}
\label{fig6}
\end{figure}
As shown in the conducted experiments the proposed receiver has a clear advantage over the conventional receiver in noise conditions, this is mostly due to the fact that the proposed LSTM demodulator takes advantage of the statistics of the generating speech signal. We prove this point by limiting the memory of the network to only one time step, as in theory we can map the FM signal back to audio with almost no memory. Though we were able to reconstruct the audio for noise free case with rather small reconstruction error of SNR 17.76 dB, however for low SNR conditions, of 0 dB amplitude noise, reconstruction was not possible without using memory, and the demodulation failed. This experiment shows that indeed in order for quality reconstruction to take place under noise conditions the statistics of the generating speech signal must be accounted for. 
Next we compare the reconstruction quality of the proposed LSTM demodulator and conventional demodulator, in the presence of phase noise. This is done by adding AWGN noise both to the modulating speech signal and to the frequency modulated signal separately. We add phase noise by adding AWGN with SNR of 0 dB with respect to the speech signal and conduct experiments employing FM demodulation for various levels of amplitude AWGN. 
Fig. 7 shows MSE of the speech reconstruction comparison between the proposed LSTM demodulator and conventional
demodulator for phase noise of 0 dB with respect to the speech signal and for various levels of AWGN amplitude noise.
Fig. 8 shows PESQ score of the speech reconstruction comparison between the proposed LSTM demodulator and conventional
demodulator for phase noise of 0 dB and for various levels of AWGN amplitude noise.
It can be seen that the LSTM demodulator outperforms the conventional demodulator in the case of both amplitude and phase noise, thus creating an end-to-end radio receiver that can overcome both communication channel disturbances and acoustical disturbances, modeled as phase noise.
  

%
%
%
\section{Conclusion}
We have presented a new approach to decode FM transmission of audio speech signals based on bidirectional stacked LSTM.
In this approach we utilize the statistics of the information message, more specifically long and short time-scale temporal structure in speech. As a result the proposed receiver has a clear advantage over the conventional receiver as it yields much higher reconstruction quality and can overcome both distortions in the information message and distortions in the transmission channel.
With the availability of sufficient computation power, which became practical with the appearance of powerful graphical processing units (GPU) and corresponding software, the proposed receiver can be used as an extremely robust radio receiver.

\bibliography{main}{}

\begin{thebibliography}{1}

\bibitem{digidemod1}
M.~Amini and E.~Balarastaghi.
\newblock {Universal Neural Network Demodulator for Software Defined Radio}.
\newblock {\em Journal of Machine Learning and Computing}, 1(3):305--310, 2011.

\bibitem{digidemod3}
Meng Fan and Lenan Wu.
\newblock {Demodulator based on deep belief networks in communication system}.
\newblock {\em International Conference on Communication, Control, Computing
  and Electronics Engineering}, 2017.

\bibitem{fan}
Yuchen Fan, Yao Qian, Fenglong Xie, and Frank K.Soong.
\newblock {TTS Synthesis with Bidirectional LSTM based Recurrent Neural
  Networks}.
\newblock {\em Interspeech}, page 1964–1968, 2014.

\bibitem{speech3}
Tobias Goehringa, Federico Bolnerb, Jessica~J.M. Monaghana, Bas van Dijkc,
  Andrzej Zarowskid, and Stefan Bleecka.
\newblock {Speech enhancement based on neural networks improves speech
  intelligibility in noise for cochlear implant users}.
\newblock {\em Hearing Research}, 344:183 – 194, 2017.

\bibitem{graves13}
Alex Graves.
\newblock Generating sequences with recurrent neural networks.
\newblock {\em CoRR}, abs/1308.0850, 2013.

\bibitem{stack}
Alex Graves, Abdel rahman Mohamed, and Geoffrey Hinton.
\newblock {Speech recognition with deep recurrent neural networks}.
\newblock {\em Acoustics, Speech and Signal Processing (ICASSP)}, 2013.

\bibitem{matlab}
Indranil Hatai and Indrajit Chakrabarti.
\newblock A new high-performance digital fm modulator and demodulator for
  software-defined radio and its fpga implementation.
\newblock {\em Int. J. Reconfig. Comp.}, 2011:342532:1--342532:10, 2011.

\bibitem{lstm}
Sepp Hochreiter and Jürgen Schmidhuber.
\newblock {Long Short-Term Memory}.
\newblock {\em Neural Computation}, 9(8):1735--1780, 1997.

\bibitem{speechlstm}
Morten Kolbek, Zheng-Hua Tan, and Jesper Jensen.
\newblock {Speech enhancement using Long Short-Term Memory based recurrent
  Neural Networks for noise robust Speaker Verification}.
\newblock {\em Spoken Language Technology Workshop}, 2016.

\bibitem{speech2}
Anurag Kumar and Dinei Florencio.
\newblock {Speech Enhancement In Multiple-Noise Conditions using Deep Neural
  Networks}.
\newblock {\em arXiv}, 2016.

\bibitem{stack2}
Xiangang Li and Xihong Wu.
\newblock {Constructing long short-term memory based deep recurrent neural
  networks for large vocabulary speech recognition}.
\newblock {\em Acoustics, Speech and Signal Processing (ICASSP)}, 2015.

\bibitem{gruv}
Aran Nayebi and Matt Vitelli.
\newblock {GRUV: Algorithmic Music Generation using Recurrent Neural Networks}.

\bibitem{digidemod4}
Mursel Onder, Aydın Akan, and Hakan Dogan.
\newblock {Neural network based receiver design for Software Defined Radio over
  unknown channels}.
\newblock {\em Electrical and Electronics Engineering}, 2013.

\bibitem{digidemod2}
Mursel Onder, Aydın Akan, and Hakan Dogan.
\newblock {Advanced neural network receiver design to combat multiple channel
  impairments}.
\newblock {\em Turkish Journal of Electrical Engineering and Computer
  Sciences}, 1(24):33066 – 3077, 2016.

\bibitem{Pascanu:2013:DTR:3042817.3043083}
Razvan Pascanu, Tomas Mikolov, and Yoshua Bengio.
\newblock On the difficulty of training recurrent neural networks.
\newblock In {\em Proceedings of the 30th International Conference on
  International Conference on Machine Learning - Volume 28}, ICML'13, pages
  III--1310--III--1318. JMLR.org, 2013.

\bibitem{anademod}
Kamyar Rohani and Michael T.Manry.
\newblock {The Design of Multi-Layer Perceptrons Using Building Blocks}.
\newblock {\em IJCNN-91-Seattle International Joint Conference on Neural
  Networks}, 1991.

\bibitem{Rumelhart:1988:LRB:65669.104451}
David~E. Rumelhart, Geoffrey~E. Hinton, and Ronald~J. Williams.
\newblock Neurocomputing: Foundations of research.
\newblock chapter Learning Representations by Back-propagating Errors, pages
  696--699. MIT Press, Cambridge, MA, USA, 1988.

\bibitem{sutskever}
Ilya Sutskever, Oriol Vinyals, and Quoc~V. Le.
\newblock {Sequence to Sequence Learning with Neural Networks}.

\bibitem{rmsprop}
T.~Tieleman and G.~Hinton.
\newblock {Lecture 6.5---RmsProp: Divide the gradient by a running average of
  its recent magnitude}.
\newblock COURSERA: Neural Networks for Machine Learning, 2012.

\bibitem{prob}
Richard~E. Turner and Maneesh Sahani.
\newblock {Demodulation as Probabilistic Inference}.
\newblock {\em IEEE Transactions on Audio, Speech, and Language Processing},
  19(8):2398 -- 2411, 2011.

\bibitem{commbook}
Gregory~W. Wornell.
\newblock {\em Efficient Symbol-Spreading Strategies for Wireless
  Communication}.
\newblock Research Laboratory of Electronics, Massachusetts Institute of
  Technology, Cambridge , Massachusetts, 1994.

\bibitem{speech1}
Yong Xu, Jun Du, Li-Rong Dai, and Chin-Hui Lee.
\newblock {A Regression Approach to Speech Enhancement Based on Deep Neural
  Networks}.
\newblock {\em IEEE/ACM Transactions on Audio, Speech, and Language
  Processing}, 23(1), 2015.

\bibitem{zen}
Heiga Zen and Hasim Sak.
\newblock {Unidirectional Long Short-Term Memory Recurrent Neural Network with
  Recurrent Output Layer for Low-Latency Speech Synthesis}.

\bibitem{timit}
Garofolo, J. S. and Lamel, L. F. and Fisher, W. M. and Fiscus, J. G. and Pallett, D. S. and Dahlgren, N. L.
\newblock{DARPA TIMIT Acoustic Phonetic Continuous Speech Corpus}, 1993


\bibitem{brnn}
Mike Schuster and Kuldip K. Paliwal and A. General,
\newblock  {Bidirectional recurrent neural networks},
\newblock {\em IEEE Transactions on Signal Processing}, 23(1), 2015.
    journal = {IEEE Transactions on Signal Processing},1997


\bibitem{dropout}
Nitish Srivastava and Geoffrey Hinton and Alex Krizhevsky and Ilya Sutskever and Ruslan Salakhutdinov,
\newblock  {Dropout: A Simple Way to Prevent Neural Networks from Overfitting},
\newblock {\em Journal of Machine Learning Research}, 15, 2014

\bibitem{pesq}
A.~W. Rix, J.~G. Beerends, M.~P. Hollier, and A.~P. Hekstra.
\newblock Perceptual evaluation of speech quality (pesq)-a new method for
  speech quality assessment of telephone networks and codecs.
\newblock In {\em Proceedings of the Acoustics, Speech, and Signal Processing,
  200. On IEEE International Conference - Volume 02}, ICASSP '01, pages
  749--752, Washington, DC, USA, 2001. IEEE Computer Society.

\end{thebibliography}
\bibliographystyle{plain}

\end{document}